# Exploring the Potential of Feature Density in Estimating Machine Learning Classifier Performance with Application to Cyberbullying Detection

Juuso Eronen[1], Michal Ptaszynski[1], Fumito Masui[1], Gniewosz Leliwa[2] and Michal Wroczynski[2]
[1]Kitami Institute of Technology, Japan
[2]Samurai Labs, Poland
eronen.juuso@gmail.com, {ptaszynski, f-masui}@cs.kitami-it.ac.jp, {gniewosz.leliwa, michal.wroczynski}@samurailabs.ai

## Abstract

In this research, we analyze the potential of Feature Density (FD) as a way to comparatively estimate machine learning (ML) classifier performance prior to training. The goal of the study is to aid in solving the problem of resource-intensive training of ML models which is becoming a serious issue due to continuously increasing dataset sizes and the ever rising popularity of Deep Neural Networks (DNN). The issue of constantly increasing demands for more powerful computational resources is also affecting the environment, as training large-scale ML models are causing alarmingly-growing amounts of $CO_2$ emissions. Our approach is to optimize the resource-intensive training of ML models for Natural Language Processing to reduce the number of required experiments iterations. We expand on previous attempts on improving classifier training efficiency with FD while also providing an insight to the effectiveness of various linguistically-backed feature preprocessing methods for dialog classification, specifically cyberbullying detection.

## 1 Introduction

One of the challenges in machine learning (ML) has always been estimating how well different classification algorithms will perform with a given dataset. Although there are classifiers that tend to be highly effective on a variety of different problems, they might be easily outperformed by others on a dataset specific scale. As it is difficult to identify a classifier that would perform best with every kind of dataset [Michie *et al.*, 1995], it comes down to the user (researcher, or ML practitioner) to determine experimentally, which classifier could be appropriate based on their knowledge of the field and previous experiences.

A common way when estimating the performance of different classifiers is to select a variety of possible classifiers to experiment on and train them using cross-validation to aid in getting the best possible average estimations of their performances. With a sufficiently small dataset and using a computationally efficient algorithm, this approach works very well. Even though it is possible to get accurate estimations of the classifier performance this way, it is multiple times more costly.

Previously, there have been some attempts to estimate the performance of a ML model before any training. One proposal to this problem is using meta-learning and training a model using dataset characteristics to estimate classifier performance [Gama and Brazdil, 1995]. Another approach is extrapolating results from small datasets to simulate the performance using larger datasets [Basavanhally *et al.*, 2010].

The importance of resolving this issue comes not only from the increased computational requirements, but also from its environmental effect. This is directly caused by the increased popularity of the fields of Artificial Intelligence (AI) and ML. Training classifiers on large datasets is both time consuming and computationally intensive while leaving behind a noticeable carbon footprint [Strubell *et al.*, 2019]. To move towards greener AI [Schwartz *et al.*, 2019], it is necessary to inspect the core of ML methods and find potential points of improvement. In order to save computational power and reduce emissions, it would be useful to roughly estimate classifier performance prior to training.

The ability to estimate classifier performance before the training would also have important practical implications. In dialog agent applications, one of the areas where the need for this is becoming more urgent is in forum moderation, specifically the detection of harmful and abusive behaviour observed online, known as cyberbullying (CB). The number of CB cases has been constantly growing since the increase of the popularity of Social Networking Services (SNS) [Hinduja and Patchin, 2010; Ptaszynski and Masui, 2018]. The consequences of unattended cases of online abuse are known to be serious, leading the victims to self mutilation, or even suicides, or on the opposite, to attacking their offenders in revenge. Being able to roughly estimate which classifier settings can be rejected, would make the process of implementation of automatic cyberbullying detection for various languages and social networking platforms more efficient.

To contribute to that, we conduct an in-depth analysis of the effectiveness of FD proposed previously by [Ptaszynski *et al.*, 2017] to comparatively estimate the performance of different classifiers before training. We also analyze the effectiveness of various linguistically-backed feature preprocessing methods, including lemmas, Named Entity Recognition (NER) and dependency information-based features, with an application to automatic cyberbullying detection.

## 2 Previous Research
### 2.1 Classifier Performance Estimation
[Gama and Brazdil, 1995] proposed that classifier performance could be estimated by training a regression model based on meta-level characteristics of a dataset. The characteristics used included simple measures like number of examples and number of attributes, statistical measures like standard deviation ratio and various information based measures like class entropy. These measures are defined in the STATLOG project [King *et al.*, 1995].

This meta-learning approach was taken further by [Bensusan and Kalousis, 2001] who introduced the Landmarking method, using learners themselves to characterize the datasets. This means using computationally non-demanding classifiers, like Naive Bayes (NB), to obtain important insights about the datasets. The method outperformed the previous characterization method and had moderate success in ranking learners.

Later, [Blachnik, 2017] improved on the Landmarking method by proposing the use of information from instance selection methods as landmarks. These instance selection methods are most commonly used for cleaning the dataset reducing it size by removing redundant information. They discovered that the relation between the original and reduced datasets can be used as a landmark to lower the error rates when predicting classifier performance.

Another approach to predicting classifier performance is to extrapolate results from a smaller dataset to simulate the performance of a larger dataset. [Basavanhally *et al.*, 2010] attempted to predict classifier performance in the field of computer aided diagnostics, where data is very often limited in quantity. Their experiments showed that using a repeated random sampling method on small datasets to make predictions on a larger set tended to have high error rates and should not be generalized as holding true when large amounts of data become available. Later, [Basavanhally *et al.*, 2015] improved this method by utilizing it together with cross-validation sampling strategy, which resulted in lower error rates.

In the field of NLP, [Johnson *et al.*, 2018] applied the extrapolation method to document classification using the fastText classifier. They discovered that biased power law model with binomial weights works as a good baseline extrapolation model for NLP tasks.

Instead of concentrating on meta information of the dataset or performance simulation, our research directly targets feature engineering and the relation between the available feature space and classifier performance. This novel method that can be utilized together with the existing methods to better estimate the performance of different classifiers.

### 2.2 Feature Density
The concept of Feature Density (FD) was introduced by [Ptaszynski *et al.*, 2017] based on the notion of Lexical Density [Ure, 1971] from linguistics. It is a score representing an estimated measure of content per lexical units for a given corpus, calculated as the number of all unique words divided by the number of all words in the corpus. The score is called Feature Density as it also includes other features, like parts-of-speech or dependency information, in addition to words.

In this research, after calculating FD for all applied dataset preprocessing methods we calculated Pearson's correlation coefficient ($\rho$-value) between dataset generalization (FD) and classifier results (F-scores). If ideal ranges of FD can be identified, or FD has a positive or negative correlation with classifier performance, it could be useful in comparatively estimating the performance of various classifiers. For example, [Ptaszynski *et al.*, 2017] showed that CNNs benefit from higher FD while other classifiers' score was usually higher when using lower FD datasets. This suggests that it could be possible to improve the performance of CNNs by increasing the FD of the applied dataset, while other classifiers could achieve higher scores by lowering FD [Ptaszynski *et al.*, 2017].

In practice, we attempt to estimate what feature engineering methods can achieve the highest performance for different models in different languages. The method lets us ignore redundant feature sets for a particular classifier or language and only keep the ones with the highest performance potential without actually training any models.

### 2.3 Linguistically-backed Preprocessing
Almost without exception, the word embeddings are learned from pure tokens (words) or lemmas (unconjugated forms of words). This is also the case with the recently popularized pre-trained language models like BERT [Devlin *et al.*, 2018]. To the best of our knowledge, embeddings backed with linguistic information have not yet been researched extensively, with only a handful of related work attempting to explore the subject [Levy and Goldberg, 2014; Komninos and Manandhar, 2016; Cotterell and Schütze, 2019].

To further investigate the potential of capturing deeper relations between lexical items and structures and to filter out redundant information, we propose to preserve the morphological, syntactic and other types of information by adding linguistic information to the pure tokens or lemmas. This means, for example, including parts-of-speech or dependency information within the used lexical features. These combinations would then be used to train the word embeddings. The method could be later applied to the pre-training of huge language models to possibly improve their performance. The preprocessing methods are described in-depth in section 3.2.

## 3 Dataset and Learners
### 3.1 Dataset
We tested the concept of FD on the Kaggle Formspring Dataset for Cyberbullying Detection [Reynolds *et al.*, 2011]. However, the original dataset had a problem of being annotated by laypeople, whereas it has been pointed out before that datasets for topics such as online harassment and cyberbullying should be annotated by experts [Ptaszynski and Masui, 2018]. Therefore in our research we applied the version of the dataset after re-annotation with the help of highly trained data annotators with sufficient psychological background to assure high quality of annotations [Ptaszynski *et*

Table 1: Statistics of the dataset after improved annotation.

| Element type | Value |
|---|---|
| Number of samples | 12,772 |
| Number of CB samples | 913 |
| Number of non-CB samples | 11,859 |
| Number of all tokens | 301,198 |
| Number of unique tokens | 18,394 |
| Avg. length (chars) of a post (Q+A) | 12.1 |
| Avg. length (words) of a post (Q+A) | 23.6 |
| Avg. length (chars) of a question | 61.6 |
| Avg. length (words) of a question | 12 |
| Avg. length (chars) of a answer | 58.5 |
| Avg. length (words) of a answer | 11.5 |
| Avg. length (chars) of a CB post | 12.1 |
| Avg. length (words) of a CB post | 22.9 |
| Avg. length (chars) of a non-CB post | 13.9 |
| Avg. length (words) of a non-CB post | 24.7 |

*al.*, 2018]. Cyberbullying is a phenomenon observed in many SNS. It is defined as using online means of communication to harass and/or humiliate individuals. This can include slurry comments about someone's looks or personality or spreading sensitive or false information about individuals. This problem has existed throughout the time of communication via Internet between people but has grown extensively with the advent of communication devices that can be used on-the-go like smartphones and tablets. Users' realization of the anonymity of online communications is one of the factors that make this activity attractive for bullies since they rarely face consequences of their improper behavior [Bull, 2010]. The problem has been growing with the popularity of SNS.

Table 1 reports some key statistics of the current annotation of the dataset. The dataset contains approximately 300 thousand of tokens. There were no visible differences in length between the posted questions and answers (approx. 12 words). On the other hand, the harmful (CB) samples were usually slightly shorter than the non-harmful (non-CB) samples (approx. 23 vs. 25 words). The number of harmful samples was small, amounting to 7%, which roughly reflects the amount of profanity on SNS [Ptaszynski and Masui, 2018].

### 3.2 Preprocessing

In order to train the linguistically-backed embeddings, we first preprocessed the dataset in various ways, similarly to [Ptaszynski *et al.*, 2017]. This was done to verify the correlation between the classification results and Feature Density (FD) and to verify the performance of various versions of the proposed linguistically-backed embeddings. The preprocessing was done using spaCy NLP toolkit (https://spacy.io/). After assembling combinations from the listed preprocessing types, we ended up with a total of 68 possible preprocessing methods for the experiments. The FDs for all separate preprocessing types used in this research were shown in Table 2.

- **Tokenization:** includes words, punctuation marks, etc. separated by spaces (later: TOK).
- **Lemmatization:** like the above but with generic (dictionary) forms of words ("lemmas") (later: LEM).
- **Parts of speech (separate):** parts of speech information is added in the form of separate features (later: POSS).
- **Parts of speech (combined):** parts of speech information is merged with other applied features (later: POS).
- **Named Entity Recognition (without replacement):** information on what named entities (private name of a person, organization, numericals, etc.) appear in the sentence are added to the applied word (later: NER).
- **Named Entity Recognition (with replacement):** same as above but information replaces the applied word (later: NERR).
- **Dependency structure:** noun- and verb-phrases with syntactic relations between them (later: DEP).
- **Chunking:** like above but without dependency relations ("chunks", later: CHNK).
- **Stopword filtering:** redundant words are filtered out using spaCy's stopword list for English (later: STOP)
- **Filtering of non-alphabetics:** non-alphabetic characters are filtered out (later: ALPHA)

### 3.3 Feature Extraction

We generated a Bag-of-Words language model from each of the 68 processed dataset versions. This resulted in separate models for each of the datasets (Bag-of-Words, Bag-of-Lemmas, Bag-of-POS, etc.). Next, we applied a weighting scheme, term frequency with inverse document frequency or $tf * idf$.

When training a Convolutional Neural Network model, the embeddings were trained as a part of the network for all of the described datasets. Similarly to other classifiers, we trained a separate model for each of the 68 datasets (Word/token Embeddings, Lemmas Embeddings, POS Embeddings, Chunks Embeddings, etc.). The embeddings were trained as part of the network using Keras' embedding layer with random initial weights, meaning no pretraining was used.

### 3.4 Classification

We used two variants of Support Vector Machine [Cortes and Vapnik, 1995], **linear-SVM** and linear-SVM with **SGD** optimizer. We also used two different solvers for Logistic Regression (LR), **Newton** and **L-BFGS**. We also used both **AdaBoost** [Freund and Schapire, 1997] and **XGBoost** [Chen and Guestrin, 2016]. Other classifiers applied include **Random Forest** [Breiman, 2001], **kNN**, **NaïveBayes**, Multilayer Perceptron (**MLP**) And Convolutional Neural Network (**CNN**).

In this experiment MLP refers to a network using regular dense layers. We applied an MLP implementation with Rectified Linear Units (ReLU) as a neuron activation function [Hinton *et al.*, 2012] and one hidden layer with dropout regularization which reduces overfitting and improves generalization by randomly dropping out some of the hidden units during training [Hinton *et al.*, 2012].

We applied a CNN implementation with Rectified Linear Units (ReLU) as a neuron activation function, and max pooling [Scherer *et al.*, 2010], which applies a max filter to non-overlying sub-parts of the input to reduce dimensionality and in effect correct overfitting. We also applied dropout regularization on penultimate layer. We applied two versions of CNN. First, with one hidden convolutional layer containing 128 units. The second version consisted of two hidden convolutional layers, containing 128 feature maps each, with 4x4 size of patch and 2x2 max-pooling, and Adaptive Moment Estimation (Adam), a variant of Stochastic Gradient Descent [LeCun et al., 2012].

Table 2: Feature Density of preprocessing types.

| Preprocessing type | Uniq.1grams | All1grams | FD |
|---|---|---|---|
| POS | 18 | 357616 | .0001 |
| POSALPHA | 18 | 357616 | .0001 |
| POSSTOP | 18 | 194606 | .0001 |
| POSSTOPALPHA | 17 | 129076 | .0001 |
| LEMPOSSALPHA | 17875 | 579664 | .0308 |
| LEMPOSS | 21238 | 660653 | .0321 |
| TOKPOSSALPHA | 21737 | 579624 | .0375 |
| TOKPOSS | 25122 | 660612 | .038 |
| LEMNERALPHA | 14815 | 289868 | .0511 |
| LEMNERR | 17327 | 309124 | .0561 |
| CHNKNERRALPHA | 12293 | 215096 | .0572 |
| LEMNERRALPHA | 17877 | 305481 | .0585 |
| CHNKNERALPHA | 14007 | 228146 | .0614 |
| LEMALPHA | 17860 | 289868 | .0616 |
| LEMPOSSSTOP | 20948 | 334870 | .0626 |
| TOKNERRALPHA | 18595 | 289828 | .0642 |
| CHNKALPHA | 13991 | 215096 | .065 |
| LEMNER | 21239 | 325173 | .0653 |
| LEMPOSSSTOPALPHA | 17554 | 258103 | .068 |
| TOKNERR | 21119 | 309084 | .0683 |
| LEM | 21222 | 308434 | .0688 |
| TOKNERALPHA | 21737 | 305441 | .0712 |
| TOKPOSSSTOP | 24472 | 334869 | .0731 |
| LEMPOS | 26232 | 357657 | .0733 |
| TOKALPHA | 21722 | 289828 | .0749 |
| LEMPOSALPHA | 22206 | 289868 | .0766 |
| TOKNER | 25121 | 325132 | .0773 |
| TOK | 25106 | 308393 | .0814 |
| TOKPOSSSTOPALPHA | 21037 | 258103 | .0815 |
| TOKPOS | 31121 | 357616 | .087 |
| TOKPOSALPHA | 27013 | 289828 | .0932 |
| LEMNERRSTOPALPHA | 14509 | 129076 | .1124 |
| LEMNERRSTOP | 17047 | 146549 | .1163 |
| LEMNERSTOPALPHA | 17557 | 142289 | .1234 |
| CHNKNERR | 33025 | 262529 | .1258 |
| LEMNERRSTOPALPHA | 20950 | 160269 | .1307 |
| LEMPOSSTOP | 25669 | 194674 | .1319 |
| LEMSTOPALPHA | 17540 | 129076 | .1359 |
| TOKNERRSTOPALPHA | 17911 | 129076 | .1387 |
| CHNKNER | 38044 | 272581 | .1396 |
| TOKNERRSTOP | 20480 | 146549 | .1397 |
| LEMSTOP | 20933 | 145866 | .1435 |
| CHNKNERSTOPALPHA | 13356 | 92782 | .144 |
| CHNKNERRSTOPALPHA | 11656 | 80896 | .1441 |
| CHNK | 38029 | 261990 | .1452 |
| TOKNERSTOPALPHA | 21037 | 142289 | .1478 |
| TOKNERSTOP | 24471 | 160268 | .1527 |
| TOKPOSSTOP | 30040 | 194673 | .1543 |
| TOKSTOPALPHA | 21022 | 129076 | .1629 |
| CHNKSTOPALPHA | 13340 | 80896 | .1649 |
| LEMPOSSTOPALPHA | 21626 | 129076 | .1675 |
| TOKSTOP | 24456 | 145865 | .1677 |
| TOKPOSSTOPALPHA | 25925 | 129076 | .2009 |
| CHNKNERRSTOP | 32452 | 126357 | .2568 |
| CHNKNERSTOP | 37462 | 135357 | .2768 |
| CHNKSTOP | 37447 | 125824 | .2976 |
| DEPNERALPHA | 95404 | 240302 | .397 |
| DEPNERRALPHA | 94928 | 215096 | .4413 |
| DEPALPHA | 95386 | 215096 | .4435 |
| DEPNER | 143197 | 321835 | .4449 |
| DEPNERSTOPALPHA | 47159 | 104940 | .4494 |
| DEPNERR | 141479 | 308704 | .4583 |
| DEP | 143179 | 308704 | .4638 |
| DEPNERSTOP | 94539 | 184130 | .5134 |
| DEPNERRSTOP | 92730 | 172086 | .5389 |
| DEPSTOP | 94521 | 172086 | .5493 |
| DEPNERRSTOPALPHA | 46552 | 80896 | .5755 |
| DEPSTOPALPHA | 47141 | 80896 | .5827 |

Table 3: Classifiers with best F1, preprocessing type and Pearson's correlation coefficient for FD and F1.

| Classifier | Best F1 | Best PP type | $\rho$(F1, FD) |
|---|---|---|---|
| SGD SVM | .798 | TOKPOS | -.8239 |
| MLP | .7958 | TOK | -.8599 |
| Linear SVM | .7941 | TOKPOSSTOP | -.834 |
| L-BFGS LR | .7932 | TOKSTOP | -.8024 |
| Newton LR | .7915 | TOKNERSTOP | -.8097 |
| RandomForest | .7582 | TOKSTOP | -.7873 |
| XGBoost | .7523 | LEMSTOP | -.8303 |
| CNN1 | .7406 | DEPSTOP | .1633 |
| CNN2 | .7357 | LEMPOSS | .0951 |
| AdaBoost | .7356 | TOKSTOP | -.7362 |
| NaiveBayes | .7165 | TOK | -.7531 |
| KNN | .6711 | TOKPOSSSTOPALPHA | -.7116 |

## 4 Experiments

### 4.1 Setup

The preprocessed dataset provides 68 separate datasets and the experiment was performed once for each preprocessing type. Each of the classifiers (sect. 3.4) were tested on each version of the dataset in a 10-fold cross validation procedure. This gives us an opportunity to evaluate how effective different preprocessing methods are for each classifier. As the dataset was not balanced, we oversampled the minority class using Synthetic Minority Over-sampling Technique (SMOTE) [Chawla et al., 2002]. The preprocessing methods represent a wide range of Feature Densities, which can be used to evaluate the correlation with classifier performance.

### 4.2 Effect of Feature Density

We analyzed the correlation of Feature Density with each of the classifiers using the proposed preprocessing methods. The results are represented in Table 5. As the results for using only parts-of-speech tags, which had the lowest FD by far, were extremely low (close to a coinflip). Thus, we can say that POS tags alone do not contain enough information to successfully classify the entries.

After excluding the preprocessing methods that only used POS tags, we can see that all classifiers, except CNNs have a strong negative correlation with Feature Density. So these classifiers seem to have a weaker performance if a lot of linguistic information is added, and the best results being usually within the range of .05 to .15 FD depending on the classifier. This range includes 38 of the 68 preprocessing methods (Table 2), meaning that the total training time could be reduced by around 40-50%. This can be seen from, for example, the highest performing classifier, SVM with SGD optimizer (Figure 1), where the maximum classifier performance starts high at around .05 and slowly falls until .14 after which there is a noticeable drop. The performance only falls further as the FD rises.

For CNNs however, there was a very weak positive or no correlation between FD and the classifier performance, with the higher FD datasets performing equally or even slightly better when comparing to the low FD datasets. Taking a look at one layer CNN's performance, which was better than the CNN with two layers, we can see from Figure 1 that the maximum performance starts at a moderate level and stays more stable throughout the whole range of feature densities. The most potential ranges of FD are between .05 to .1 and after .45. The potential training time reduction seems to be similar, around 40-50% The reduction in training time could be especially important when considering demanding models like Neural Networks.

The results suggest that for non-CNN classifiers there is no need to consider preprocessings with a high FD, such as chunking or dependencies, as they had a considerably lower performance. The performance seems to start falling rapidly at around $FD = .15$ with most of the classifiers. For CNNs, high performances were recorded on both low and high FDs. This means that there is potential in the higher FD preprocessing types, namely, dependencies for CNNs.

The reason for CNNs relatively low performance could be explained by the relatively small size of the dataset, especially when considering the amount of actual cyberbullying entries, as adding even a second layer to the network already caused a loss of the most valuable features and ended up degrading performance. With such small amount of data, it doesn't seem useful to train deep learning models to solve the classification problem. Still, the dependency based features are showing potential with CNNs. With a considerably larger dataset and more computational power, it could be possible to outperform other classifiers and the usage of tokens with dependency based features when using deep learning.

The experiments show that changing Feature Density in moderate amount can yield good results when using other classifiers than CNNs. However, excessive changes to either too low or too high always showed diminishing results. The treshold was in all cases approximately between 50% and 200% of the original density (TOK), most optimal FDs only slightly varying with each classifier. The exception being Random Forest [Breiman, 2001], which showed a clear spike at around .12 FD. As the usage of high Feature Density datasets showed potential with CNNs, their usage needs to be confirmed in future research. Also, more exact ideal feature densities need to be confirmed for each classifier using datasets of different sizes and fields to make a more accurate ranking of classifiers by FD possible.

### 4.3 Analysis of Linguistically-backed Preprocessing

From the results it can be seen that most of the classifiers scored highest on pure tokens. CNNs also performed quite well on the dependency-based preprocessings. Using lemmas usually got slightly lower scores than tokens probably due to information loss. Chunking got low performance overall and was clearly outperformed by dependency-based features in CNNs. Using only POS tags achieved very low performance and thus it should be only used as a supplement to other methods.

Stopword filtering seemed to be the one of the most effective preprocessing techniques for traditional classifiers, which can be seen from Table 3 as it was used in the majority of the highest scores. The problem with stopwords was that the scores fluctuated a lot, having both low and very high scores and scoring high mostly with Logistic Regression and all of the tree based classifiers. An important thing to note is that the preprocessing method had extremely polarized performance with CNNs, scoring either very high or low. Overall, stopwords yielded the most top scores of any preprocessing method considering all the classifiers.

Another very effective preprocessing method was Parts of Speech merging (POS), which achieved high performance overall when added to TOK or LEM. The method also got the highest scores with multiple classifiers, especially SVMs. Adding parts-of-speech information to the respective words achieved a higher score than using them as a separate feature. This keeps the information directly connected to the word itself, which seems a better option when preserving information.

Using Named Entity Recognition reduced the classifier performance most of the time, only achieving a high score with one classifier, Newton-LR. The performance of using NER seemed clearly inferior compared to stopwords or POS information. Replacing words with their NER information seems to cause too much information loss and reduces the performance when comparing to plain tokens. Attaching NER information to the respective words did not improve the performance in most cases but still performed better than replacement. These results are different to [Ptaszynski *et al.*, 2017], who noticed that NER helped most of the times for cyberbullying (CB) detection in Japanese. This could come from the fact that CB is differently realized in those languages. In Japan, revealing victim's personal information, or "doxxing" is known to be one of the most often used form of bullying, thus NER, which can pin-point information such as address or phone number often help in classification, while this is not the case in English.

Filtering out non-alphabetic characters also reduced the classifier performance most of the time and also got a high score with only one classifier, kNN, which was the weakest classifier overall. Non-alphabetic tokens seem to carry useful information, at least in the context of cyberbullying detection, as removing them reduced the performance comparing to plain tokens due to information loss.

Trying to generalize the feature set ended up lowering the results in most cases with the exception of the very high scores of stopword filtering using traditional classifiers. This would mean that the stopword filter sometimes succeeded in removing noise and outliers from the dataset while other generalization methods ended up cutting useful information. Adding information to tokens could be useful in some scenarios as was shown with parts-of-speech tags and using dependency information with CNNs, although using NER was not so successful. Any kind of generalization attempt resulted in a lower performance with CNNs, which shows their ability to assemble more complex patterns from tokens and relations

Table 4: Approximate power usage of the training processes. Non-neural classifiers: i9 7920X, 163W. Neural classifiers: GTX 1080ti, 250W. Expecting 100% power usage.

| Classifier | Runtime (s) | Power usage (Wh) | Best F1 |
|---|---|---|---|
| SGD SVM | 176.26 | 79.81 | .798 |
| MLP | 53845.89 | 37392.98 | .7958 |
| Linear SVM | 1543.06 | 698.67 | .7941 |
| L-BFGS LR | 321.6 | 145.61 | .7932 |
| Newton LR | 249.74 | 113.08 | .7915 |
| Random Forest | 3982.49 | 1803.18 | .7582 |
| XGBoost | 17917.74 | 8112.76 | .7523 |
| CNN1 | 62361.45 | 43306.56 | .7406 |
| CNN2 | 62054.46 | 43093.37 | .7357 |
| AdaBoost | 10425.4 | 4720.39 | .7356 |
| Naive Bayes | 97.54 | 44.16 | .7165 |
| KNN | 556.44 | 251.94 | .6711 |

that are unusable by other classifiers.

An interesting discovery is that using raw tokens only rarely resulted in the best performance considering the proposed feature sets. This can be seen from Tables 3 and 5. This proves the effectiveness of using linguistics-based feature engineering instead of directly using words as features. Also, the performance of one-layer CNN increased significantly when using linguistic embeddings, from 0.659 (TOK) F-score to 0.741 (DEPSTOP). The high scores of dependency-based feature sets indicate that structural information could be important.

In order to compare the usage of linguistic preprocessing to modern text classifiers, we fine-tuned RoBERTa [Liu *et al.*, 2019] on the dataset. This showed an F-score of 0.797, which is similar to the highest scores by other models using our method. Actually, the best score by SGD SVM is 0.798 which is slightly higher. It is fascinating that a simple method like SVM can outperform a complex modern text classifier when using the right feature set. This shows that traditional, more simple models should not be underestimated as with correct preparations, they can achieve a similar performance as state-of-the art models and require much less computational power. Possibly, the performance of pretrained language models like RoBERTa could also be increased by feature engineering and applying embeddings with linguistic information. This needs to be explored further in future research.

### 4.4 Environmental Effect

If the weaker feature sets were to be left out, the power savings are approximately 35Wh calculated from Table 4 for training the SGD SVM classifier, which is not very much. But the classifier was very power efficient to train to begin. A more impressive result can be seen with CNN, where the power savings are approximately 21kWh, which is considerably more compared to SVM.

In order demonstrate the environmental effect of the method, we will look at the CNN model and its power savings (21kWh). According to European Environmental Agency (EEA) [1], the average $CO_2$ emissions of electricity generation was 275 $g$ $CO_2e$/kWh in 2019. Thus the greenhouse gases emitted during the training of CNN could be estimated to be 5.8 $kg$ $CO_2e$. For comparison, the average new passenger car in the European Union in 2019, according to EEA, emits around 122 $g$ $CO_2e$ per kilometer driven. So when training a simple CNN model, if we calculate the feature densities and leaving out the weaker feature sets before training, we could save as much as driving a new car for almost 50 kilometers in emissions.

Instead of having to run all of the experiments, it could be useful to first discard the FD ranges of the overall weakest feature sets. Then running a small subset of the experiments with a set interval between preprocessing type feature densities, look for the FD range with a high performance and iterate around it by running more experiments with similar feature densities in order to find the maximum performance.

## 5 Conclusions

In this paper we presented our research on Feature Density and linguistically-backed preprocessing methods, applied in dialog classification and cyberbullying detection. Both concepts are relatively novel to the field. We studied the effect of FD in reducing the number of required experiments iterations and analyzed the usage of different linguistically-backed preprocessing methods in the context of CB detection.

The results indicate that for non-CNN classifiers, there is an ideal Feature Density that slightly differs between each classifier. This can be taken into account in future experiments in order to save time and computational power when running experiments. For CNNs however, there is almost no correlation between FD and classifier and thus the higher FD datasets should also be considered when trying to achieve the best performance.

Using plain tokens to keep the original words and their forms and reducing noise with stopword filtering yielded the best results in general. With some classifiers, adding extra information in the form of POS tags also proved useful. For convolutional neural networks, using dependency based information showed potential and their effect needs to be confirmed in future research.

Although the environmental effect of the method does not seem very significant here, one has to keep in mind that the tested models were quite simple. Assuming that the method would work with other datasets and more resource intensive classifiers, the savings could be very significant. It could be useful to only run a subset of the experiments and iterate around the most probable performance peak in order to find the maximum performance.

In the near future we will also confirm the potential of linguistically-backed preprocessing and Feature Density for other applications and languages. The research further suggests that adding linguistic preprocessing can improve the performance of classifiers, which needs to be also confirmed on current state of the art language models.

## References

[Basavanhally *et al.*, 2010] A. Basavanhally, S. Doyle, and A. Madabhushi. Predicting classifier performance with a small training set: Applications to computer-aided diagnosis and prognosis. In *2010 IEEE International Symposium*

---
[1] https://www.eea.europa.eu/

Table 5: F1 for all preprocessings & classifiers; best classifier for each dataset in **bold**; best preprocessing for each underlined

| | LBFGS LR | Newton LR | Linear SVM | SGD SVM | KNN | NaiveBayes | RandomForest | AdaBoost | XGBoost | MLP | CNN1 | CNN2 |
|---|---|---|---|---|---|---|---|---|---|---|---|---|
| CHNK | 0.727 | 0.726 | 0.718 | **0.736** | 0.57 | 0.674 | 0.613 | 0.649 | 0.667 | 0.724 | 0.657 | 0.666 |
| CHNKNERR | 0.688 | 0.695 | 0.702 | 0.699 | 0.58 | 0.653 | 0.603 | 0.608 | 0.642 | **0.704** | 0.645 | 0.662 |
| CHNKNERRALPHA | 0.66 | 0.663 | 0.651 | 0.657 | 0.603 | 0.626 | 0.616 | 0.599 | 0.653 | **0.674** | 0.566 | 0.6 |
| CHNKNERRSTOP | 0.686 | 0.684 | 0.684 | **0.694** | 0.577 | 0.629 | 0.635 | 0.621 | 0.652 | 0.693 | 0.402 | 0.344 |
| CHNKNERRSTOPALPHA | 0.618 | 0.617 | 0.591 | 0.607 | 0.404 | 0.598 | 0.62 | 0.582 | **0.648** | 0.623 | 0.451 | 0.34 |
| CHNKNER | 0.718 | 0.723 | 0.721 | **0.737** | 0.582 | 0.669 | 0.603 | 0.63 | 0.673 | 0.722 | 0.654 | 0.642 |
| CHNKNERALPHA | 0.675 | 0.676 | 0.663 | 0.663 | 0.599 | 0.641 | 0.618 | 0.609 | 0.649 | **0.684** | 0.557 | 0.614 |
| CHNKNERSTOP | **0.724** | **0.724** | 0.715 | **0.724** | 0.582 | 0.663 | 0.635 | 0.652 | 0.679 | 0.72 | 0.501 | 0.298 |
| CHNKNERSTOPALPHA | 0.666 | 0.661 | 0.644 | **0.668** | 0.386 | 0.615 | 0.659 | 0.625 | 0.656 | 0.647 | 0.431 | 0.406 |
| CHNKALPHA | 0.684 | 0.681 | 0.669 | 0.683 | 0.607 | 0.643 | 0.647 | 0.616 | 0.676 | **0.695** | 0.587 | 0.583 |
| CHNKSTOP | 0.722 | 0.721 | 0.711 | **0.723** | 0.577 | 0.67 | 0.667 | 0.648 | 0.679 | 0.715 | 0.386 | 0.342 |
| CHNKSTOPALPHA | 0.629 | 0.637 | 0.606 | 0.619 | 0.395 | 0.608 | 0.649 | 0.654 | **0.664** | 0.628 | 0.455 | 0.374 |
| DEP | 0.617 | 0.619 | 0.568 | 0.587 | 0.243 | 0.617 | 0.536 | 0.566 | 0.598 | 0.594 | 0.682 | **0.694** |
| DEPNERR | 0.61 | 0.614 | 0.571 | 0.587 | 0.241 | 0.611 | 0.533 | 0.562 | 0.596 | 0.595 | 0.67 | **0.695** |
| DEPNERRALPHA | 0.606 | 0.605 | 0.589 | 0.602 | 0.312 | 0.596 | 0.537 | 0.556 | 0.595 | 0.593 | 0.585 | **0.622** |
| DEPNERRSTOP | 0.602 | 0.599 | 0.564 | 0.568 | 0.273 | 0.615 | 0.543 | 0.572 | 0.6 | 0.578 | **0.726** | 0.702 |
| DEPNERRSTOPALPHA | 0.584 | 0.584 | 0.56 | 0.581 | 0.386 | 0.599 | 0.544 | 0.561 | 0.595 | 0.574 | 0.583 | **0.619** |
| DEPNER | 0.624 | 0.621 | 0.574 | 0.585 | 0.242 | 0.611 | 0.528 | 0.564 | 0.595 | 0.592 | 0.686 | **0.692** |
| DEPNERALPHA | 0.585 | 0.589 | 0.561 | 0.579 | 0.213 | 0.607 | 0.578 | 0.497 | 0.593 | 0.603 | 0.606 | **0.623** |
| DEPNERSTOP | 0.611 | 0.602 | 0.564 | 0.576 | 0.274 | 0.604 | 0.527 | 0.563 | 0.604 | 0.577 | **0.725** | 0.708 |
| DEPNERSTOPALPHA | 0.535 | 0.531 | 0.523 | 0.523 | 0.297 | 0.543 | 0.563 | 0.422 | 0.576 | 0.564 | 0.63 | **0.632** |
| DEPALPHA | 0.609 | 0.612 | 0.588 | 0.601 | 0.314 | 0.6 | 0.545 | 0.552 | 0.604 | 0.598 | 0.606 | **0.62** |
| DEPSTOP | 0.606 | 0.595 | 0.562 | 0.571 | 0.276 | 0.616 | 0.544 | 0.576 | 0.603 | 0.584 | **0.741** | 0.648 |
| DEPSTOPALPHA | 0.586 | 0.587 | 0.564 | 0.588 | 0.388 | 0.594 | 0.539 | 0.568 | 0.595 | 0.578 | **0.629** | 0.625 |
| LEM | 0.781 | 0.786 | 0.784 | **0.79** | 0.634 | 0.715 | 0.724 | 0.72 | 0.744 | 0.786 | 0.67 | 0.665 |
| LEMNERR | 0.74 | 0.737 | 0.742 | 0.74 | 0.601 | 0.692 | 0.697 | 0.683 | 0.724 | **0.749** | 0.658 | 0.663 |
| LEMNERRALPHA | 0.729 | 0.728 | 0.725 | 0.725 | 0.614 | 0.685 | 0.699 | 0.68 | 0.71 | **0.74** | 0.645 | 0.652 |
| LEMNERRSTOP | 0.737 | 0.734 | 0.726 | 0.732 | 0.609 | 0.682 | 0.727 | 0.69 | 0.72 | **0.741** | 0.371 | 0.364 |
| LEMNERRSTOPALPHA | 0.732 | 0.732 | 0.714 | 0.727 | 0.624 | 0.674 | 0.723 | 0.682 | 0.704 | **0.737** | 0.372 | 0.348 |
| LEMPOSS | 0.764 | 0.765 | 0.769 | 0.767 | 0.564 | 0.713 | 0.658 | 0.679 | 0.717 | **0.773** | 0.662 | 0.736 |
| LEMPOSSALPHA | **0.76** | 0.758 | 0.753 | 0.758 | 0.406 | 0.705 | 0.669 | 0.674 | 0.712 | 0.756 | 0.603 | 0.715 |
| LEMPOSSSTOP | 0.763 | 0.766 | 0.767 | **0.774** | 0.566 | 0.709 | 0.706 | 0.691 | 0.72 | 0.773 | 0.683 | 0.725 |
| LEMPOSSSTOPALPHA | 0.762 | **0.766** | 0.748 | 0.765 | 0.49 | 0.702 | 0.713 | 0.681 | 0.714 | 0.757 | 0.593 | 0.716 |
| LEMNER | 0.784 | 0.782 | 0.787 | **0.792** | 0.631 | 0.71 | 0.716 | 0.72 | 0.742 | 0.78 | 0.68 | 0.613 |
| LEMNERALPHA | 0.763 | 0.764 | 0.765 | 0.767 | 0.637 | 0.699 | 0.71 | 0.707 | 0.742 | **0.768** | 0.662 | 0.671 |
| LEMNERSTOP | 0.782 | 0.783 | 0.782 | **0.792** | 0.634 | 0.706 | 0.745 | 0.725 | 0.742 | 0.78 | 0.429 | 0.378 |
| LEMNERSTOPALPHA | **0.77** | 0.767 | 0.752 | 0.767 | 0.64 | 0.693 | 0.739 | 0.716 | 0.738 | 0.768 | 0.46 | 0.414 |
| LEMPOS | 0.778 | 0.778 | 0.788 | **0.79** | 0.517 | 0.711 | 0.663 | 0.727 | 0.741 | 0.783 | 0.665 | 0.64 |
| LEMPOSALPHA | 0.768 | 0.772 | 0.772 | 0.768 | 0.522 | 0.7 | 0.654 | 0.713 | 0.727 | **0.775** | 0.664 | 0.695 |
| LEMPOSSTOP | 0.78 | 0.781 | **0.788** | **0.788** | 0.642 | 0.708 | 0.708 | 0.721 | 0.735 | 0.783 | 0.715 | 0.707 |
| LEMPOSSTOPALPHA | 0.77 | 0.769 | 0.766 | 0.768 | 0.669 | 0.696 | 0.718 | 0.722 | 0.73 | **0.778** | 0.669 | 0.698 |
| LEMALPHA | 0.755 | 0.764 | 0.745 | **0.765** | 0.294 | 0.703 | 0.718 | 0.705 | 0.748 | 0.754 | 0.61 | 0.651 |
| LEMSTOP | 0.787 | 0.786 | 0.784 | **0.791** | 0.641 | 0.713 | 0.754 | 0.732 | 0.752 | 0.789 | 0.403 | 0.327 |
| LEMSTOPALPHA | 0.772 | 0.766 | 0.766 | **0.773** | 0.357 | 0.702 | 0.747 | 0.712 | 0.745 | 0.764 | 0.377 | 0.329 |
| POSS | 0.487 | 0.487 | 0.488 | 0.491 | 0.522 | 0.498 | **0.556** | 0.509 | 0.555 | 0.488 | 0.54 | 0.536 |
| POSSALPHA | 0.488 | 0.486 | 0.498 | 0.498 | 0.526 | 0.498 | **0.552** | 0.518 | 0.549 | 0.493 | 0.538 | 0.534 |
| POSSSTOP | 0.477 | 0.477 | 0.471 | 0.467 | 0.518 | 0.486 | **0.54** | 0.496 | 0.533 | 0.484 | 0.431 | 0.434 |
| POSSSTOPALPHA | 0.469 | 0.47 | 0.471 | 0.465 | 0.517 | 0.478 | **0.525** | 0.484 | 0.511 | 0.491 | 0.428 | 0.484 |
| TOK | 0.793 | 0.788 | 0.793 | **0.796** | 0.632 | 0.711 | 0.728 | 0.748 | **0.796** | 0.659 | 0.661 |  |
| TOKNERR | 0.741 | 0.744 | 0.737 | 0.743 | 0.6 | 0.696 | 0.688 | 0.671 | 0.719 | **0.749** | 0.655 | 0.631 |
| TOKNERRALPHA | 0.734 | 0.735 | 0.735 | 0.73 | 0.624 | 0.683 | 0.681 | 0.674 | 0.704 | **0.748** | 0.626 | 0.655 |
| TOKNERRSTOP | 0.736 | 0.736 | 0.728 | 0.732 | 0.609 | 0.68 | 0.73 | 0.678 | 0.71 | **0.751** | 0.406 | 0.317 |
| TOKNERRSTOPALPHA | 0.728 | 0.731 | 0.727 | 0.723 | 0.623 | 0.675 | 0.721 | 0.68 | 0.698 | **0.744** | 0.412 | 0.394 |
| TOKPOSS | 0.766 | 0.768 | 0.767 | **0.783** | 0.549 | 0.715 | 0.648 | 0.671 | 0.715 | 0.773 | 0.686 | 0.729 |
| TOKPOSSALPHA | 0.765 | 0.761 | 0.763 | 0.767 | 0.378 | 0.709 | 0.662 | 0.656 | 0.709 | **0.769** | 0.643 | 0.658 |
| TOKPOSSSTOP | 0.763 | 0.765 | 0.767 | **0.773** | 0.563 | 0.704 | 0.703 | 0.684 | 0.724 | 0.771 | 0.675 | 0.722 |
| TOKPOSSSTOPALPHA | 0.774 | 0.773 | 0.774 | 0.771 | 0.671 | 0.694 | 0.722 | 0.713 | 0.73 | **0.779** | 0.68 | 0.698 |
| TOKNER | **0.789** | 0.785 | 0.788 | **0.789** | 0.609 | 0.708 | 0.703 | 0.722 | 0.745 | 0.784 | 0.684 | 0.68 |
| TOKNERALPHA | 0.768 | 0.771 | 0.763 | 0.776 | 0.628 | 0.696 | 0.701 | 0.705 | 0.746 | 0.775 | 0.649 | 0.648 |
| TOKNERSTOP | 0.785 | **0.791** | 0.79 | 0.79 | 0.635 | 0.703 | 0.732 | 0.721 | 0.743 | 0.79 | 0.444 | 0.367 |
| TOKNERSTOPALPHA | 0.773 | 0.771 | 0.762 | **0.774** | 0.646 | 0.691 | 0.737 | 0.704 | 0.74 | 0.771 | 0.371 | 0.379 |
| TOKPOS | 0.781 | 0.783 | 0.791 | **0.798** | 0.565 | 0.713 | 0.656 | 0.72 | 0.739 | 0.787 | 0.626 | 0.705 |
| TOKPOSALPHA | 0.775 | 0.775 | 0.778 | **0.784** | 0.576 | 0.699 | 0.653 | 0.705 | 0.731 | 0.783 | 0.633 | 0.698 |
| TOKPOSSTOP | 0.786 | 0.783 | **0.794** | 0.792 | 0.645 | 0.7 | 0.711 | 0.733 | 0.739 | 0.789 | 0.706 | 0.691 |
| TOKPOSSTOPALPHA | 0.759 | **0.766** | 0.756 | 0.762 | 0.458 | 0.696 | 0.706 | 0.679 | 0.674 | 0.601 | 0.734 | 0.718 |
| TOKALPHA | 0.768 | 0.768 | 0.757 | **0.773** | 0.271 | 0.705 | 0.721 | 0.705 | 0.742 | 0.756 | 0.643 | 0.652 |
| TOKSTOP | 0.793 | 0.79 | 0.784 | **0.794** | 0.644 | 0.708 | 0.758 | 0.736 | 0.749 | 0.787 | 0.355 | 0.321 |
| TOKSTOPALPHA | 0.775 | **0.776** | 0.766 | **0.776** | 0.342 | 0.7 | 0.745 | 0.714 | 0.744 | 0.765 | 0.452 | 0.425 |

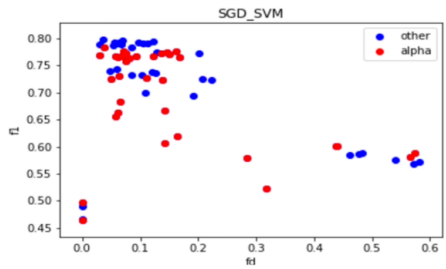
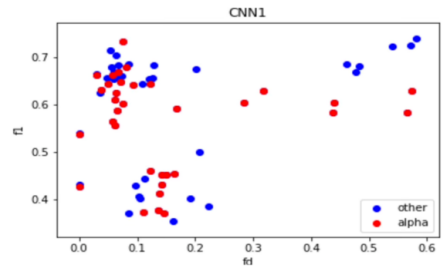

(a) Alphabetic filtering (red) vs others (blue)　　(b) Alphabetic filtering (red) vs others (blue)

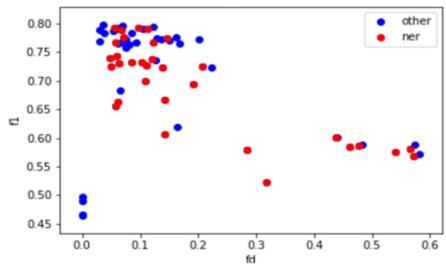
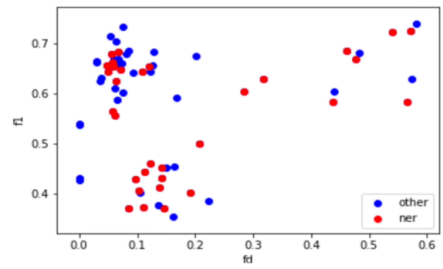

(c) NER (red) vs others (blue)　　(d) NER (red) vs others (blue)

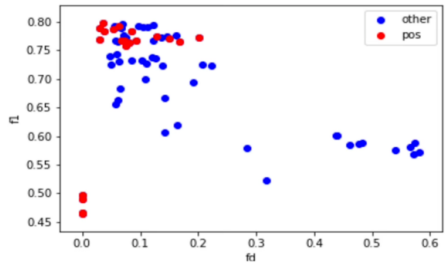
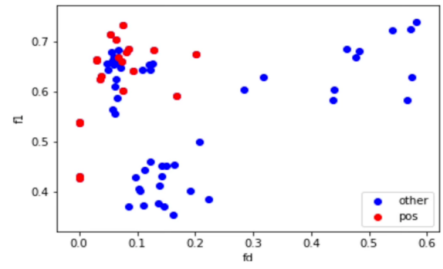

(e) POS (red) vs others (blue)　　(f) POS (red) vs others (blue)

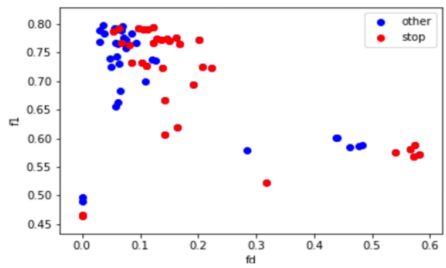
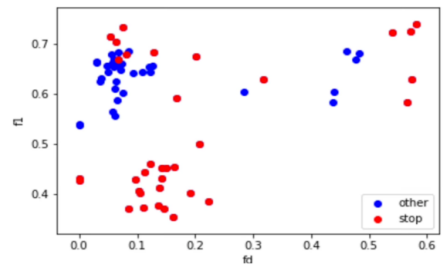

(g) Stopword filtering (red) vs others (blue)　　(h) Stopword filtering (red) vs others (blue)

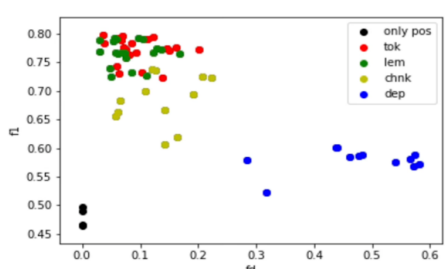
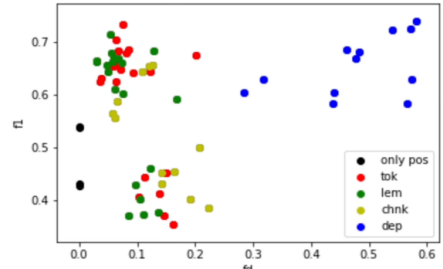

(i) TOK (red), LEM (green), CHNK (yellow), DEP (blue)　　(j) TOK (red), LEM (green), CHNK (yellow), DEP (blue)

Figure 1: FD & F1 score for SGD SVM (left) and CNN1 (right)